\documentclass[runningheads]{llncs}

\usepackage{eccv}

\usepackage{eccvabbrv}

\usepackage{graphicx}
\usepackage{booktabs}

\usepackage[accsupp]{axessibility}  

\usepackage[pagebackref,breaklinks,colorlinks,citecolor=eccvblue]{hyperref}

\usepackage{orcidlink}

\usepackage{multirow}
\usepackage{threeparttable}
\usepackage{colortbl}

\usepackage{tabularx}
\usepackage{ltablex}
\usepackage{array}

\usepackage{array}   
\usepackage{tikz}
\usetikzlibrary{calc}
\newcommand{\tikzmark}[1]{%
  \tikz[overlay,remember picture,baseline] \node (#1) {};
}
\usepackage{makecell}
\usepackage{enumitem}
\usepackage{wrapfig}

\begin{document}

\title{CAT-SAM: Conditional Tuning for Few-Shot Adaptation of Segment Anything Model} 

\titlerunning{CAT-SAM}

\author{Aoran Xiao\inst{1*}\orcidlink{0000-0002-2956-0613} \and
Weihao Xuan\inst{2,3*}\orcidlink{0009-0001-4271-9035} \and
Heli Qi\inst{4}\orcidlink{0000-0001-9512-7140} \and
Yun Xing\inst{1}\orcidlink{0000-0001-9839-0120} \and
Ruijie Ren\inst{5}\orcidlink{0009-0000-3444-4150} \and \\ 
Xiaoqin Zhang\inst{6}\orcidlink{0000-0003-0958-7285} \and
Ling Shao\inst{7}\orcidlink{0000-0002-8264-6117} \and
Shijian Lu\inst{1\dagger}\orcidlink{0000-0002-6766-2506} 
}

\authorrunning{A. Xiao and W. Xuan et al.}

\institute{Nanyang Technological University, Singapore \and
The University of Tokyo, Japan \and
RIKEN AIP, Japan \and
Nara Institute of Science and Technology, Japan \and 
Waseda University, Japan \and 
Zhejiang University of Technology, China \and
UCAS-Terminus AI Lab, University of Chinese Academy of Sciences, China}

\maketitle

\renewcommand{\thefootnote}{\textsuperscript{}}
\begin{abstract}
\footnotetext{* Co-first authors with equal contributions. $^\dagger$ Corresponding author.}
\footnotetext{Project page: \url{https://xiaoaoran.github.io/projects/CAT-SAM}}
    The Segment Anything Model (SAM) has demonstrated remarkable zero-shot capability and flexible geometric prompting in general image segmentation. However, it often struggles in domains that are either sparsely represented or lie outside its training distribution, such as aerial, medical, and non-RGB images. Recent efforts have predominantly focused on adapting SAM to these domains using fully supervised methods, which necessitate large amounts of annotated training data and pose practical challenges in data collection. This paper presents CAT-SAM, a ConditionAl Tuning network that explores \textit{few-shot} adaptation of SAM toward various challenging downstream domains in a data-efficient manner. The core design is a \textit{prompt bridge} structure that enables \textit{decoder-conditioned joint tuning} of the heavyweight image encoder and the lightweight mask decoder. The bridging maps the domain-specific features of the mask decoder to the image encoder, fostering synergic adaptation of both components with mutual benefits with few-shot target samples only, ultimately leading to superior segmentation in various downstream tasks. We develop two CAT-SAM variants that adopt two tuning strategies for the image encoder: one injecting learnable prompt tokens in the input space and the other inserting lightweight adapter networks. Extensive experiments over 11 downstream tasks show that CAT-SAM achieves superior segmentation consistently even under the very challenging one-shot adaptation setup.
  
\end{abstract}

\section{Introduction}\label{sec:intro}
Accurate image segmentation is a fundamental computer vision task that plays a pivotal role in various applications such as robotics, autonomous driving, healthcare, earth observation, etc. The recently developed Segment Anything Model (SAM)~\cite{kirillov2023segment}, trained with 1.1 billion masks, has emerged as a momentous leap forward in image segmentation. By taking geometric prompts with points, boxes, or masks as input, SAM demonstrates remarkable zero-shot capability for general image segmentation, as well as great potential for fine-grained mask segmentation in different downstream tasks across a variety of contexts.

\begin{figure}[t]
    \centering
    \includegraphics[width=0.7\linewidth]{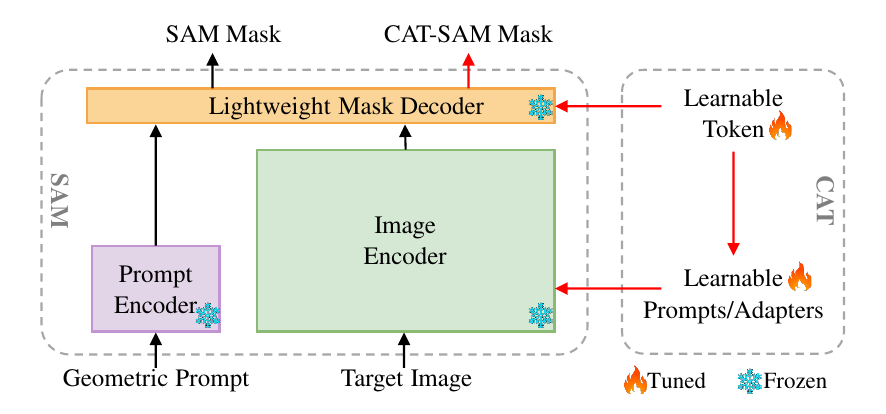}
    \caption{The proposed CAT-SAM performs ConditionAl joint Tuning (CAT) to establish communication between SAM's heavyweight image encoder and lightweight mask decoder. This enables synergistic adaptation of the two network components, mitigating tuning imbalances and improving few-shot SAM adaptation.}
    \label{fig:motivation}
\end{figure}
 
On the other hand, SAM often fails to generate high-quality predictions when dealing with domains that are either sparsely represented or lie outside its training distribution~\cite{chen2023sam,ji2023segment}, such as challenging RGB images including aerial, medical, and intricate structural images, as well as non-RGB images like X-ray, Sonar, Synthetic Aperture Radar (SAR) images, etc.
This greatly undermines the versatility and applicability of SAM as a foundational segmentation model while handling various downstream tasks. 
Several studies attempted to address this issue by fine-tuning SAM with a large number of target samples with mask annotations~\cite{ke2023segment,ma2024segment}. 
However, they require extensive masked images for each specialized field, posing a formidable task with poor scalability. How to effectively adapt SAM toward various downstream tasks in a data-efficient manner has become an essential challenge in the segmentation field.

We exploit \textit{few-shot} target samples for effective and efficient adaptation of SAM toward various downstream tasks. 
Inspired by the recent advancements in parameter-efficient tuning for foundation models \cite{bommasani2021opportunities,gu2023systematic}, we freeze the entire SAM and lightly expand its image encoder and mask decoder with a tiny amount of learnable parameters. This expansion preserves SAM's zero-shot capabilities and flexibility while capturing domain-specific features that are essential for the segmentation of downstream data. However, SAM’s image encoder is much larger than its mask decoder, which directly leads to tuning imbalance and further suboptimal adaptation, especially under the limited supervision of few-shot target samples.

To address this challenge, we design CAT-SAM, a novel ConditionAl Tuning network for effective and data-efficient adaptation of SAM. The key idea is to exploit the lightweight decoder tuning to guide the heavyweight encoder tuning and formulate \textit{decoder-conditioned joint tuning} of both as depicted in ~\cref{fig:motivation}. To this end, we design a \textit{prompt bridge} structure that maps the domain-specific features from the mask decoder to the image encoder. With the collaboration between the two partners, CAT-SAM enables synergistic adaptation of both which mitigates the tuning imbalance and improves the few-shot SAM adaptation effectively. In addition, the prompt bridge can be seamlessly embedded into two representative tuning approaches, including prompt tuning~\cite{li2021prefix,lester2021power,jia2022visual} that introduces learnable prompt tokens into the input space and adapters~\cite{rebuffi2017learning,rebuffi2018efficient, houlsby2019parameter,liu2023explicit} that leverage lightweight adaptive networks. This directly leads to two CAT-SAM variants, namely, CAT-SAM-T and CAT-SAM-A. Extensive evaluations over 11 diverse and challenging downstream tasks show that both variants achieve superior adaptation and segmentation with few-shot target samples only.

The major contributions of this work can be summarized in three aspects. \textit{First}, we propose CAT-SAM, a conditional tuning network that enables effective and data-efficient adaptation of SAM toward various challenging downstream domains. We design \textit{prompt bridge} within CAT-SAM, a decoder-conditioned joint tuning structure that enables synergistic and data-efficient adaptation of the heavyweight image encoder and the lightweight mask decoder effectively. \textit{Second}, we develop two CAT-SAM variants by embedding the prompt bridge into two representative tuning strategies, one introducing learnable prompt tokens in the input space and the other inserting lightweight adapter networks. \textit{Third}, extensive experiments over 11 diverse segmentation datasets show that CAT-SAM achieves superior image segmentation consistently even under the challenging one-shot setup.

\section{Related Work}\label{sec.related-work}

\noindent\textbf{Parameter-Efficient Tuning} for foundation models has become increasingly crucial due to the over-size model parameters and costly one-task-one-model deployment. One popular approach involves updating only the newly added learnable parameters while keeping the model backbone frozen. These lightweight parameters are often introduced in two ways: 1) as prompt tuning~\cite{zhou2022learning, jia2022visual, khattak2023maple,gong2024coda} by adding learnable tokens to input tokens at transformer layers, 2) as adapters~\cite{rebuffi2017learning, rebuffi2018efficient, liu2023explicit, gao2023clip,xu2023side} by integrating learnable lightweight sub-networks into transformer layers. However, prior visual studies primarily focus on adapting single backbone modules~\cite{jia2022visual,liu2023explicit,chen2022adaptformer} or dual encoder backbones for vision-language tasks~\cite{khattak2023maple,zhou2022conditional}, which are suboptimal tuning solutions while applying to SAM due to its particular architecture of a heavyweight image encoder and a lightweight mask decoder. In contrast, we explore a novel conditional joint tuning strategy tailored to SAM, which facilitates its adaptation even with limited downstream data.

\noindent\textbf{SAM for Downstream Tasks.} 
The release of SAM has spurred several subsequent studies. Some concentrate on expanding semantic recognition capabilities~\cite{yang2023recognize, wang2023sam, li2023semantic, FoodSAM, zhang2023personalize}, while others aim to create more lightweight SAM variants~\cite{zhao2023fast, zhang2023faster,xiong2023efficientsam} for faster computations. Several studies explore SAM's adaptation in underperforming downstream scenarios, primarily by directly fine-tuning SAM's mask decoder~\cite{Chen_2023_ICCV, zhang2023customized,cheng2023sam,wu2023medical,song2024simada,ma2024segment}, which can severely degrade SAM's zero-shot capabilities. 
The recent HQ-SAM~\cite{ke2023segment} freezes the entire pre-trained SAM and prompt-tunes the mask decoder with a new mask head for adaptive mask prediction. 
However, these adaptation studies rely on large target datasets with a considerable number of annotated images, which are often costly to collect.
In contrast, our CAT-SAM achieves superior adaptation with only few-shot target samples, significantly reducing adaptation costs and expanding the applicability of SAM as a visual foundation model.

\section{Method}

\subsection{Preliminaries of SAM}

SAM~\cite{kirillov2023segment} handles each image with three key modules: a heavyweight \textit{image encoder} (i.e., ViT~\cite{dosovitskiy2020image} as the backbone) that extracts image embeddings, a \textit{prompt encoder} that encodes the geometric prompts (i.e., points, a box, or a coarse mask) to generate prompt embeddings, and a lightweight \textit{mask decoder} that combines the two types of embeddings to predict segmentation masks. The released SAM model is trained on a super large-scale SA-1B dataset, which consists of over 11 million images as well as 1.1 billion automatically generated masks. SAM has demonstrated remarkable zero-shot capability while dealing with various conventional natural images, as well as superb flexibility in accepting various geometric prompt inputs. On the other hand, it often struggles while facing challenging downstream domains that are often sparse or falling outside of its training data distribution, such as RGB domains of aerial, medical, and intricate structural images, as well as non-RGB domains of X-ray, SAR, and Sonar images. More details about SAM are presented in~\cite{kirillov2023segment}.

\subsection{CAT-SAM}
This subsection introduces CAT-SAM, a ConditionAl Tuning network that is designed for adapting SAM toward various under-performed downstream data. The objective is to adapt SAM efficiently with only \textit{few-shot} annotated target images, meanwhile preserving SAM's powerful zero-shot capability and geometric prompting flexibility.

CAT-SAM freezes the entire SAM and facilitates the simultaneous tuning of SAM's image encoder and mask decoder to capture representative information from few-shot target samples for effective adaptation of SAM. However, a critical issue with simultaneous tuning arises from the clear imbalance between the heavyweight image encoder (308.3 M parameters for ViT-L~\cite{dosovitskiy2020image}) and the lightweight mask decoder (4.1 M), which often leads to sub-optimal adaptation while only few-shot target samples are available. To tackle this challenge, we design \textit{decoder-conditioned joint tuning}, a novel tuning strategy that mitigates the imbalance by establishing a linkage between the encoding tuning and the decoding tuning. Specifically, we design \textit{prompt bridge}---a lightweight network that maps the domain-specific features from the mask decoder to the image encoder. The bridge design facilitates joint and balanced tuning of both network components, leading to synergistic and effective adaptation of SAM with few-shot and even a single target sample.

\begin{figure}[t]
    \centering
    \includegraphics[width=\linewidth]{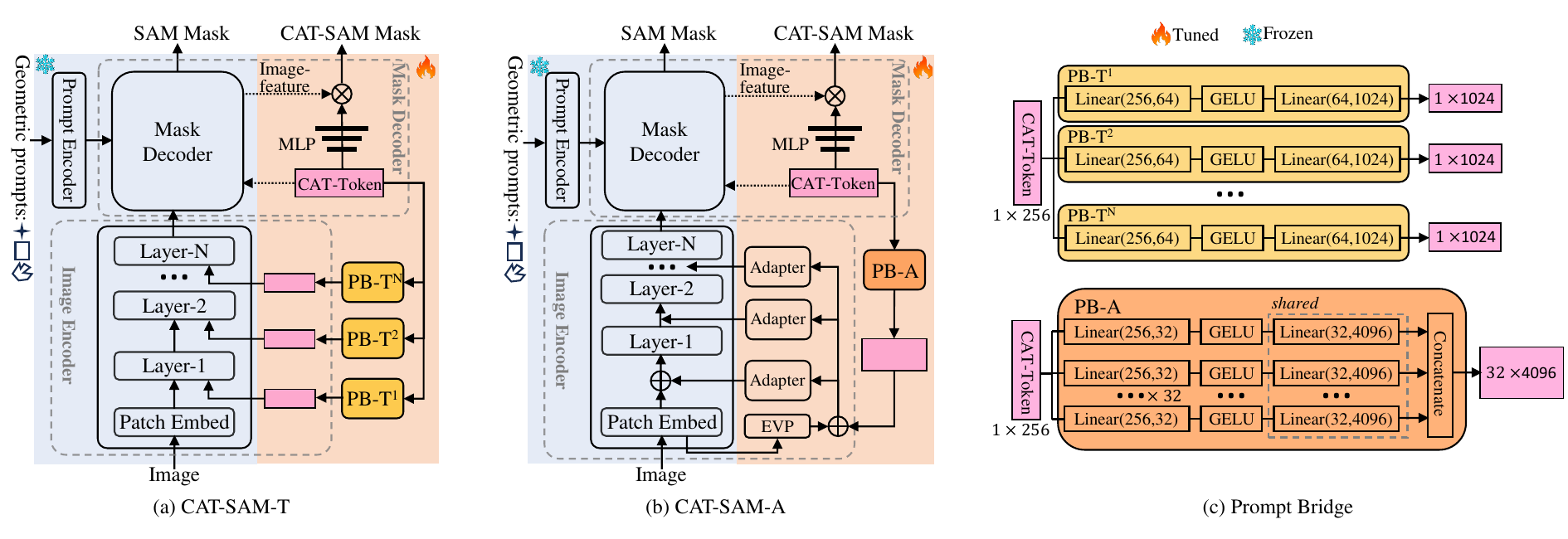}
    \caption{Overview of CAT-SAM. CAT-SAM keeps the whole SAM frozen while simultaneously tuning the image encoder and mask decoder for downstream adaptation. To address the tuning imbalance between these two network components, we introduce decoder-conditioned joint tuning through the design of Prompt Bridge structure, enabling synergetic and enhanced adaptation. We present two CAT-SAM variants: CAT-SAM-T in (a) and CAT-SAM-A in (b), achieved by integrating the prompt bridge with prompt-based and adapter-based tuning strategies for the image encoder, respectively. (c) illustrates two tailored prompt bridge structures, PB-T and PB-A.}
    \label{fig:architectures}
\end{figure}

We develop two CAT-SAM variants by integrating the proposed decoder-conditioned joint tuning into two prevalent tuning approaches, namely CAT-SAM-T and CAT-SAM-A, as illustrated  in~\cref{fig:architectures}.

\subsubsection{Tuning Image Encoder.} 
We first brief SAM's image encoder and then introduce two tuning approaches in our CAT-SAM, including prompt tuning in CAT-SAM-T and lightweight adapters in CAT-SAM-A.

\begin{itemize}[noitemsep, topsep=0pt]
    \item[$\bullet$] \textbf{SAM's Image Encoder}, denoted as $\mathcal{V}$ with $K$ transformer layers $\{\mathcal{V}_i\}^K_ {i=1}$, first splits the input image $I$ of the size $(H,W)$ into $M$ fixed-size patches, and then projects them into patch embeddings $E_0\in\mathbb{R}^{M\times d_v}$. Patch embeddings $E_{i-1}$ are input to the $i^{th}$ transformer layer $(\mathcal{V}_{i})$ and sequentially processed through $K$ transformer layers:$$[E_i] = \mathcal{V}_i([E_{i-1}]), i=1,2,...,K.$$ 
    \item[$\bullet$] \textbf{CAT-SAM-T}: As illustrated in Fig.~\ref{fig:architectures} (a), we introduce a set of learnable tokens $\{P_i\in\mathbb{R}^{d_v}\}^{K-1}_{i=0}$ in the image encoder alongside the input patch embeddings. These new learnable tokens are introduced in each transformer layer of the image encoder $\mathcal{V}$ and updated through the adaptation process: $$[E_i]=\mathcal{V}_i([E_{i-1},P_{i-1}])],i=1,2,...,K.$$
    \item[$\bullet$] \textbf{CAT-SAM-A}: As illustrated in~\cref{fig:architectures} (b), we introduce Adapters~\cite{houlsby2019parameter} to tune SAM's image encoder in CAT-SAM-A. The adapter is a lightweight sub-network that is inserted into each transformer layer, comprising a linear down-projection, a nonlinear activation function, a linear up-projection, and a residual connection. High-frequency image information is also incorporated in CAT-SAM as the input of the adapters as in~\cite{liu2023explicit}. In this process, high-frequency components $I_{hfc}$ are first extracted from the input image $I$ and then reversed to the space domain using the Fast Fourier Transform and its inverse. These components are then partitioned into small patches $I_{hfc}^p$ similar to the image patches. Subsequently, convolutional layers and linear layers project $I_{hfc}$ and the original image embeddings $E_0$ into $c$-dimensional features independently, yielding $F_{hfc}$ and $F_{pe}$, respectively. The result of element-wise addition of $F_{pe}$ and $F_{hfc}$ is then forwarded (as denoted "EVP" in~\cref{fig:architectures} (b)) to the adapter $\mathrm{Adapt}_i$ for transformer layer $i$. Finally, the output is added element-wise to the original input of the transformer layer~$i$: $$E_{i-1}=E_{i-1}+\mathrm{Adapt}_i(F_{pe}+F_{hfc}).$$
\end{itemize}

\subsubsection{Tuning Mask Decoder.}
We tune the mask decoder of SAM similarly as \cite{ke2023segment}. Specifically, we freeze the entire pre-trained decoder of SAM and introduce a learnable \textit{CAT-Token} that is concatenated with SAM's original output tokens and prompt tokens to form the input to SAM's mask decoder. 
The CAT-Token is updated during training but remains fixed during inference, the same as SAM's output tokens. After passing through two decoder layers, the updated CAT-Token is used to generate dynamic weights for a newly introduced three-layer MLP. Simultaneously, SAM's mask decoder features are fused with the image features decoded by the image encoder. Finally, the fused features and the three-layer MLP are combined through the dot product to generate the target mask.
Please refer to the appendix for more details about the decoder structures in CAT-SAM.

\subsubsection{Decoder-Conditioned Joint Tuning.} The proposed decoder-conditioned joint tuning can be seamlessly embedded via two implementations of the proposed prompt bridging ideas, i.e., PB-T and PB-A as illustrated in~\cref{fig:architectures}~(c).
\begin{itemize}[noitemsep, topsep=0pt]
    \item[$\bullet$] \textbf{CAT-SAM-T}: 
    The prompt bridge $\mathrm{PB}^T_i$ is a two-layer MLP. It projects the CAT-Token $Q$ in the mask decoder to each transformer layer $i$ in the image encoder and outputs a single learnable token replacing $P_i$ by $\tilde{P}^T_i=\mathrm{PB}^T_i(Q)$.
    These mapped tokens are directly applied for prompt tuning of the image encoder as follows: $$[E_i]=\mathcal{V}_i([E_{i-1},\tilde{P}^T_{i-1}])],i=1,2,...,K.$$ 
    \item[$\bullet$] \textbf{CAT-SAM-A}: The prompt bridge $\mathrm{PB}^A$ maps the CAT-Token $Q$ into an embedding $\tilde{P}^A=\mathrm{PB}^A(Q)$ of the same size as $F_{pe}$ and $F_{hfc}$. $\tilde{P}^A$ is then element-wise added to the EVP before being fed to every adapter as follows: $$E_{i-1}=E_{i-1}+\mathrm{Adapt}_i(F_{pe}+F_{hfc}+\tilde{P}^A).$$As illustrated in~\cref{fig:architectures} (c), PB-A employs $c$ different linear down-projection layers to generate embeddings independently. After GELU layers, these embeddings are up-projected to the dimension of $1\times M$ ($M$ is the spatial size of the patch) with a shared linear layer to mitigate the computation cost. Finally, the up-projected embeddings are concatenated to produce $\tilde{P}^A$. 
\end{itemize}

\setlength{\tabcolsep}{4.0mm}{
\begin{table}[t]
    \caption{Comparisons of trainable parameters between ViT-L~\cite{dosovitskiy2020image} based SAM and two variants of the proposed CAT-SAM.}
    \centering
    \begin{tabular}{l|c|c}
    \toprule
        Model & Trainable Parameters (M) & Trainable Params vs. SAM \\
    \midrule
        SAM~\cite{kirillov2023segment} & 312.4 & - \\
        CAT-SAM-T & 3.3  & 1.1\% \\
        CAT-SAM-A & 1.9 & 0.6\% \\
    \bottomrule
    \end{tabular}
    \label{tab:params}
\end{table}}

\cref{tab:params} compares the amount of trainable parameters between the SAM (implemented with the ViT-L~\cite{dosovitskiy2020image}) and our CAT-SAM that is built upon the SAM. It can be observed that both CAT-SAM variants introduce very limited additional parameters, yet yield significant performance improvements in various downstream segmentation tasks as to be detailed in \Cref{Sec:experiments}.

\subsection{Training and Inference}

During training, we feed target samples together with geometric prompts into CAT-SAM to generate CAT-SAM masks, under the supervision of ground-truth masks as well as a linear combination of BCE loss and dice loss~\cite{milletari2016v}. We freeze the pre-trained SAM parameters and update solely the parameters that are introduced in the above-described tuning modules.
Note we train CAT-SAM with a mixture of sampled geometric prompts, including bounding boxes, randomly selected points, and coarse masks. In addition, we create degraded coarse masks by introducing random Gaussian noise in the boundary regions of the ground-truth masks as in~\cite{ke2023segment}. Further training details can be found in Section~\ref{Subsec:implement} and the appendix.

During the inference phase, the input image is fed into SAM's image encoder along with the integrated prompt tokens in CAT-SAM-T or adapters in CAT-SAM-A, to generate adaptive image embeddings. These embeddings, combined with the prompt tokens from SAM's prompt encoder, serve as input to the mask decoder. Subsequently, the updated CAT-token and associated MLP are utilized for the target mask prediction. Lastly, we up-sample the mask to the original resolution to produce the final output.

\setlength{\tabcolsep}{5.0mm}{
\begin{table}[t]
    \centering
    \caption{We evaluate and benchmark CAT-SAM over eight challenging segmentation tasks across 11 datasets.}
    \begin{tabular}{l|l|l}
    \toprule
        Tasks & Dataset Names & Imagery \\
    \midrule
        Building segmentation & WHU~\cite{ji2018fully} & Aerial images \\ 
        Road segmentation & MA. Roads~\cite{MnihThesis} & Aerial images \\ 
        Polyp segmentation &  Kvasir~\cite{pogorelov2017kvasir} & Medical images \\ 
        Chest X-ray segmentation & JSRT~\cite{shiraishi2000development} & X-ray images \\
        Marine debris segmentation & FLS~\cite{singh2021marine} & Sonar images\\
        Ship instance segmentation & HRSID~\cite{wei2020hrsid} & SAR images \\
        Shadow segmentation & SBU-Shadow~\cite{vicente2016large} & Natural images \\ 
        \noalign{\vskip 0.3mm}
        Intricate segmentation & DIS~\cite{qin2022highly},ThinObject~\cite{liew2021deep}, & Natural images \\ 
        & HRSOD~\cite{zeng2019towards},COIFT~\cite{liew2021deep} & \\
    \bottomrule
    \end{tabular}
    \label{tab:datasets}
\end{table}}

\section{Experiments}\label{Sec:experiments}

We evaluate CAT-SAM's efficacy through comprehensive experiments involving eight segmentation tasks across 11 datasets, all hailing from challenging downstream fields that SAM struggles to address effectively.  Our experiments encompass a wide array of scenarios and tasks, spanning from varying target shots to different imagery modalities.

\subsection{Experimental Setup}
\noindent\textbf{Datasets.}
We conduct experiments on two collections of datasets listed in~\cref{tab:datasets}: 1) True-color images, including WHU~\cite{ji2018fully} for building segmentation, Kvasir~\cite{pogorelov2017kvasir} for polyp segmentation, SBU-Shadow~\cite{vicente2016large} for shadow segmentation, Massachusetts (MA.) Roads~\cite{MnihThesis} for road segmentation, as well as fine-grained segmentation datasets DIS~\cite{qin2022highly}, ThinObject~\cite{liew2021deep}, HRSOD~\cite{zeng2019towards}, and COIFT~\cite{liew2021deep}. 2) Non-RGB domains, encompassing JSRT~\cite{shiraishi2000development} for chest organ segmentation (X-ray images), HRSID~\cite{wei2020hrsid} for ship instance segmentation (high-resolution SAR images), and FLS~\cite{singh2021marine} for marine debris segmentation (Sonar images). We use official dataset splits for fair comparisons.

\noindent\textbf{Evaluation Metrics.}
For single-class datasets including WHU, Kvasir, SBU-Shadow, and MA. Roads, we employ standard mask IoU of foreground. JSRT and FLS, which have multiple classes, are evaluated using individual class IoU and their average. As for DIS, ThinObject, HRSOD, and COIFT, we follow \cite{ke2023segment} and report mIoU and the boundary metrics (mBIoU)~\cite{cheng2021boundary} for fair comparisons. For instance segmentation of HRSID, we use standard AP, AP\textsubscript{50}, and AP\textsubscript{75}. 

\noindent\textbf{Implementation Details.}\label{Subsec:implement}
Unless otherwise specified, CAT-SAM and its comparing models use ViT-L~\cite{dosovitskiy2020image} as the image encoder backbone. We train on 1 NVIDIA RTX A6000 GPU for one-shot adaptation, while 4 GPUs for multiple-shots. Following \cite{ke2023segment}, ground truth boxes serve as the default input geometric prompts during evaluation to ensure fair comparison and minimize randomness, with the exception of \cref{fig:SinglePoint} where point prompts are evaluated. Please refer to the appendix for more implementation details.

\subsection{Ablation Studies}

\begin{table}[t]
    \centering
    \caption{Ablation study of CAT-SAM-T in (a) and CAT-SAM-A in (b) for 1-shot adaptation over datasets:  "W" for WHU~\cite{ji2018fully} (on building segmentation), "K" for Kvasir~\cite{pogorelov2017kvasir} (on polyp segmentation), and "S" for SBU-Shadow~\cite{vicente2016large} (on shadow segmentation). "Enc-T." and "Enc-A." means tuning the image encoder with prompt tokens and adapters, respectively. "Dec." means tuning the mask decoder. "PB-T" and "PB-A" denote two customized prompt bridges for conditional tuning. Model 1 is the baseline SAM without adaptation.}
    \setlength{\tabcolsep}{0.4mm}
    \begin{subtable}{0.45\linewidth}
    \centering
    \begin{tabular}{l|ccc|ccc|c}
    \multicolumn{8}{c}{(a) CAT-SAM-T.} \\
    \toprule
        & Enc-T. & Dec. & PB-T & W & K & S & AVG \\
    \midrule
       1 & & & & 43.5 & 79.0 & 62.4 & 61.6\\
    \midrule
       2 & \checkmark & & & 57.8 & 80.4 & 76.0 & 71.4\\
       3 & & \checkmark & & 71.2  &73.8  &63.5    &69.5 \\
       4 & \checkmark & \checkmark & & 66.4 & 73.2 &61.8 &67.1\\
       5 & \checkmark & \checkmark & \checkmark & \textbf{86.8} & \textbf{83.4} & \textbf{78.0} & \textbf{82.7} \\
    \bottomrule
    \end{tabular}
    \end{subtable}
    \hspace{12pt}
    \begin{subtable}{.45\linewidth}
        \begin{tabular}{l|ccc|ccc|c}
        \multicolumn{8}{c}{(b) CAT-SAM-A.} \\
    \toprule
         & Enc-A. & Dec. & PB-A & W & K & S & AVG\\
    \midrule
       1 & & & & 43.5 & 79.0 & 62.4 &61.6 \\
    \midrule
       2 & \checkmark & & & 86.5 & 73.7 & 76.3 & 78.8 \\
       3 & & \checkmark & & 71.2  &73.8  &63.5 &69.5 \\
       4 & \checkmark & \checkmark & & 83.4 & 79.8 & 76.3 & 79.8 \\
       5 & \checkmark & \checkmark & \checkmark &  \textbf{88.2} & \textbf{85.4} & \textbf{81.9} & \textbf{85.2} \\
    \bottomrule
    \end{tabular}
    \end{subtable}
    \label{tab:ablation-CAT}
\end{table}

We first investigate the contributions of different tuning modules within both CAT-SAM variants to assess their impact on overall adaptation performance. \cref{tab:ablation-CAT} shows the experiments of \textit{one-shot} adaptation across three true-color image segmentation benchmarks WHU~\cite{ji2018fully}, Kvasir~\cite{pogorelov2017kvasir}, and SBU-Shadow~\cite{vicente2016large}. For each CAM-SAM variant CAM-SAM-T/CAM-SAM-A, we compare 5 models including: 1.~The original SAM without adaptation (baseline); 2.~Tuning SAM's image encoder alone with prompt tokens or adapters; 3.~Tuning SAM's mask decoder alone; 4.~Tuning SAM's image encoder and mask decoder independently; and 5.~Conditional tuning of SAM's image encoder and mask decoder with the proposed prompt bridging (i.e., the complete CAT-SAM-T and CAT-SAM-A).

We can see that the original SAM struggles to produce high-quality masks in these challenging domains, indicating its limitations as a foundational segmentation model for many downstream applications. Tuning the image encoder or the mask decoder alone in 2 and 3 enhances target segmentation, underlining the efficacy and necessity of SAM adaptation.  
However, simultaneously tuning the image encoder and mask decoder in an independent way in model 4 does not show complementary and consistent improvement. This suggests instability and imbalance in tuning between the heavyweight image encoder and the lightweight mask decoder, which could lead to suboptimal adaptation solutions. 
Differently, including the proposed prompt bridging with either PB-T or PB-A (i.e., the complete CAT-SAM-T and CAT-SAM-A) mitigates the tuning imbalance and produces substantial performance enhancement consistently.

\subsection{Comparison with the State-of-the-Art}

\noindent\textbf{CAT-SAM for One-Shot Adaptation.}
To assess CAT-SAM's effectiveness in reducing downstream training data and its adaptability with few-shot samples, we first evaluate it under the extremely challenging one-shot adaptation scenario. Given the lack of prior studies under such setup, 
we comprehensively benchmark it against closely related methods that can be broadly categorized into two groups.
The first group comprises state-of-the-art tuning methods for foundation models of different modalities, including VPT~\cite{jia2022visual} and AdaptFormer~\cite{chen2022adaptformer} for large vision foundation models, and LoRA~\cite{hu2021lora} for large language models. 
The second group comprises recent SAM-based adaptation methods which were largely designed for full-shot adaptation~\cite{ke2023segment,wu2023medical,song2024simada, zhang2023customized, ma2024segment} (\cite{wu2023self} tackles few-shot adaptation but requires 20 shots).

\setlength{\tabcolsep}{3.5mm}{
\begin{table}[t]
    \centering
    \caption{Comparison of adaptive segmentation performance on challenging \textit{true-color} datasets: WHU~\cite{ji2018fully} (on building segmentation),  Kvasir~\cite{pogorelov2017kvasir} (on Polyp segmentation), and SBU-Shadow~\cite{vicente2016large} (on shadow segmentation). The baseline is SAM without adaptation. All compared methods utilize \textbf{one-shot} sample for adaptation, except for those denoted by * which utilize 20 shots with metric numbers sourced from \cite{wu2023self}.}
    \begin{tabular}{l|ccc|c}
    \toprule
       Methods & WHU & Kvasir & SBU-Shadow & Average \\
    \midrule
      SAM~\cite{kirillov2023segment} (baseline) & 43.5 & 79.0 & 62.4 & 61.6\\
    \midrule
      VPT-\textit{shallow}~\cite{jia2022visual} & 60.8 & 79.8 & 68.7 & 69.8 \\
      VPT-\textit{deep}~\cite{jia2022visual}  & 57.8 & 80.4 & 76.0 & 71.4 \\
      AdaptFormer~\cite{chen2022adaptformer} & 83.2 &76.8 & 77.2 & 79.1 \\
      LoRA~\cite{hu2021lora} & 86.1 &77.5 & 74.4 & 79.3 \\ 
    \midrule
      Med-SA~\cite{wu2023medical} &34.5 &28.6 &22.1 & 28.4\\
      MedSAM~\cite{ma2024segment} &30.6 &29.8 &53.4 &37.9 \\
      SimAda~\cite{song2024simada} & 48.6 & 18.7 & 49.3 & 38.9\\
     SPFS-SAM*~\cite{wu2023self} (20-shots) & - & 53.4 & - & -\\
     SAMed*~\cite{zhang2023customized} (20-shots) & - & 51.7 & - & -\\
     HQ-SAM~\cite{ke2023segment} &71.2  &73.8  &63.5 &69.5 \\
    \midrule
      CAT-SAM-T (Ours) & 86.8 & 83.4 & 78.0 & 82.7 \\
      CAT-SAM-A (Ours) & \textbf{88.2} & \textbf{85.4} & \textbf{81.9} & \textbf{85.2} \\
    \bottomrule
    \end{tabular}
    \label{tab:sota-true-color}
\end{table}
}

\cref{tab:sota-true-color} presents results over WHU, Kvasir, and SBU-Shadow datasets. We can observe that all three tuning methods for foundation models exhibit clear adaptation effects. Notably, AdaptFormer and LoRA demonstrate significantly better performance than VPT-\textit{deep} and VPT-\textit{shallow}, highlighting the effectiveness of modifying the model's structure over prompt tuning in the input space during SAM adaptation.
On the other hand, most SAM-based adaptation methods, including Med-SA, SimAda, SPFS-SAM, SAMed, and MedSAM, yield much lower performance, primarily due to their reliance on many more target images which leads to overfitting while learning from one-shot target sample. HQ-SAM achieves substantial segmentation improvements but it still cannot compete with the three tuning methods, demonstrating the significance of tuning the image encoder over the mask decoder.
Finally, both CAT-SAM variants consistently outperform all compared methods, with CAT-SAM-A exhibiting a slight edge due to its network structure modifications. These experiments validate the superiority of the proposed decoder-conditioned joint tuning approach in extracting domain-specific features with few-shot samples.

\begin{table}[t]
    \caption{Comparison of state-of-the-art segmentation achieved by fully supervised learning (FSL) methods trained on full-shot data (upper part) and mask segmentation by SAM and our CAT-SAM (lower part). Evaluation follows official criteria of each FSL benchmark, reporting IoU for WHU, mIoU for Kvasir, and BER for SBU-Shadow to ensure consistency. "(1)", "(16)", and "(full)" after CAT-SAM-T/CAT-SAM-A denote tuning with 1-shot, 16-shots, and full-shots samples.}
    \setlength\tabcolsep{3.5pt}
    \centering
    \begin{tabular}{lc|lc|lc}
    \toprule
       \multicolumn{2}{c|}{WHU} & \multicolumn{2}{c|}{Kvasir} & \multicolumn{2}{c}{SBU-Shadow} \\
    \midrule
       Model & IoU & Model & mIoU & Model & BER(↓) \\
    \midrule
       STT~\cite{chen2021building} & 90.5 & CASCADE~\cite{rahman2023medical} & 87.8 & BDRAR~\cite{zhu2018bidirectional} & 3.64 \\
       BuildFormer~\cite{wang2022building} & 91.4 & PatchRefineNet~\cite{nagendra2024patchrefinenet} & 89.1 & DSDNet~\cite{zheng2019distraction} & 3.45\\
       CBR-Net~\cite{guo2022coarse} & 91.4 & DUCK-Net~\cite{dumitru2023using} & 90.5 & MTMT~\cite{chen2020multi} & 3.15 \\
    \midrule
       SAM & 43.5 & SAM & 87.3 & SAM & 13.22 \\
       CAT-SAM-T (1) & 86.8 & CAT-SAM-T (1) &90.0 & CAT-SAM-T (1) & 7.81 \\
       CAT-SAM-A (1) & 88.2  & CAT-SAM-A (1) &91.3  & CAT-SAM-A (1)  &5.27 \\
       CAT-SAM-T (16) & 89.6 & CAT-SAM-T (16) &93.1 & CAT-SAM-T (16)  &4.04 \\
       CAT-SAM-A (16) & 90.3 & CAT-SAM-A (16) & 93.6 & CAT-SAM-A (16) &3.80   \\
       CAT-SAM-T (full) & 93.3 & CAT-SAM-T (full) & \textbf{94.5} & CAT-SAM-T (full) &  2.54\\
       CAT-SAM-A (full) & \textbf{93.6} & CAT-SAM-A (full) & 94.3 & CAT-SAM-A (full) & \textbf{2.39} \\
    \bottomrule
    \end{tabular}
    \label{tab:SOTA-Fully-Supervised}
\end{table}

\noindent\textbf{Full-Shot Learning.}
While CAT-SAM is primarily designed for few-shot adaptation, it also demonstrates proficiency in full-shot adaptation scenarios. In \cref{tab:SOTA-Fully-Supervised}, we present the state-of-the-art performances of traditional fully-supervised learning (FSL) methods trained over full-shot data and compare them with SAM and our CAT-SAM (adapted with various target samples). We report official evaluation metrics for three public benchmarks, including IoU for WHU, mIoU for Kvasir, and BER~\cite{hu2018direction} for SBU-Shadow. Note the distinction in setups between the FSL segmentation and the mask segmentation: FSL focuses on traditional semantic segmentation of specific classes without prompt input while SAM and CAT-SAM involve general mask segmentation with prompt input. Despite these differences, these performances can serve as valuable references for evaluating CAT-SAM's segmentation ability.

\setlength{\tabcolsep}{0.3mm}{
\begin{table}[t]
    \centering
    \caption{Extremely fine-grained segmentation over four intricate structural image datasets. HQ-SAM and the two CAT-SAM variants are fine-tuned over full HQSeg-44K~\cite{ke2023segment}. All adopt the boxes converted from their GT masks as the box prompt input for fair comparisons.}
    \begin{tabular}{l|cc|cc|cc|cc|cc}
    \toprule
      \multirow{2}{4em}{Models} & \multicolumn{2}{c|}{DIS~\cite{qin2022highly}} & \multicolumn{2}{c|}{COIFT~\cite{liew2021deep}} & \multicolumn{2}{c|}{HRSOD~\cite{zeng2019towards}} & \multicolumn{2}{c|}{ThinObject~\cite{liew2021deep}} & \multicolumn{2}{c}{Average}\\
      & mIoU & mBIoU & mIoU & mBIoU & mIoU & mBIoU & mIoU & mBIoU & mIoU & mBIoU\\
    \midrule
      SAM~\cite{kirillov2023segment} & 62.0 & 52.8 & 92.1 & 86.5 & 90.2 &  83.1 &  73.6 & 61.8 & 79.5 & 71.1 \\
      HQ-SAM~\cite{ke2023segment} & 78.6 & 70.4 & 94.8 & 90.1 & \textbf{93.6} & 86.9 & 89.5 & 79.9 & 89.1 & 81.8 \\
      CAT-SAM-T & \textbf{84.0} & \textbf{78.1} & \textbf{95.6} & 92.0 & 93.4 & \textbf{87.6} & 94.0 & 87.9 & \textbf{91.7} & \textbf{86.4}\\
      CAT-SAM-A & 83.6 & 77.7 & \textbf{95.6} & \textbf{92.2} & 93.1 & 87.1 & \textbf{94.1} & \textbf{88.2} & 91.6 & 86.3\\
    \bottomrule
    \end{tabular}
    \label{tab:fine-seg-sota}
\end{table}}

With full target training data, both CAT-SAM variants outperform state-of-the-art FSL methods consistently. Notably, the two CAT-SAM variants achieve impressive target segmentation under the challenging one-shot setup with a single target sample. With 16 shots of target samples, their performance is even on par with the FSL methods that are trained by using full target training data. These experiments underscore the superiority of effective transfer of broad knowledge from the foundation model SAM as compared with training from scratch with FSL. It also demonstrates the remarkable efficacy of CAT-SAM in substantially expanding SAM's applicability across various downstream tasks.

We conducted further experiments to compare CAT-SAM against HQ-SAM \cite{ke2023segment}, a state-of-the-art SAM-based adaptation solution for intricate structural image segmentation. For fair comparisons, we follow the evaluation protocol of HQ-SAM and tune CAT-SAM on the complete HQSeg-44K dataset, consisting of 44,320 meticulously annotated image masks.
\cref{tab:fine-seg-sota} shows experiments across four exceptionally fine-grained segmentation datasets including DIS~\cite{qin2022highly}, ThinObject~\cite{liew2021deep}, HRSOD~\cite{zeng2019towards}, and COIFT~\cite{liew2021deep} (in mIoU and the boundary metrics (mBIoU)~\cite{cheng2021boundary}). We can see that CAT-SAM-T achieves slightly better performance than CAT-SAM-A, while both clearly outperform HQ-SAM and the original SAM. These experiments further demonstrate the efficacy of our proposed conditional tuning in adapting SAM toward challenging downstream tasks. Visual qualitative comparisons are provided in the appendix.

\subsection{CAT-SAM for Non-RGB Domains}

We also explore CAT-SAM's performance across more challenging image modalities characterized by larger domain discrepancies. 
Specifically, we assess its efficacy on non-RGB domains, including JSRT~\cite{shiraishi2000development} for chest organ segmentation using X-ray images and FLS~\cite{singh2021marine} for marine debris segmentation with Sonar images, and HRSID~\cite{wei2020hrsid} for ship instance segmentation with SAR images. For adaptation, we randomly select either one or sixteen target images for 1-shot and 16-shot scenarios, respectively, while employing the entire training set for full-shot adaptation. These datasets represent novel image distributions as compared with SAM's training data. 
Note we train instances of all classes together while evaluating them by classes. Since no images in FLS contain objects of all 11 classes, we only report results of 16-shot and full-shot adaptation for FLS.  

\setlength{\tabcolsep}{1.5mm}{
\begin{table}[t]
    \centering
    \caption{Few-shot adaptation with varying number of training samples for \textit{non-RGB} datasets, including FLS~\cite{singh2021marine} with Sonar images for marine debris segmentation, JSRT~\cite{shiraishi2000development} for chest X-ray segmentation, and HRSID~\cite{wei2020hrsid} for ship instance segmentation with SAR images. "N/A" denotes that 1-shot adaptation over FLS is unattainable due to the absence of images containing objects from all 11 classes.}
    \begin{tabular}{l|l|cc|c|c|ccc}
    \toprule
        \#Samples & Methods & \multicolumn{3}{c|}{JSRT~\cite{shiraishi2000development}} & FLS~\cite{singh2021marine} & \multicolumn{3}{c}{HRSID~\cite{wei2020hrsid}} \\
    \cmidrule{3-9}
        & & Lungs & Heart & mIoU & mIoU & AP & AP\textsubscript{50} & AP\textsubscript{75}\\
    \midrule
       None & SAM~\cite{kirillov2023segment} & 85.0 & 71.9 & 78.5 & 69.7 & 38.2 & 86.6 & 26.8 \\
    \midrule
         \multirow{2}{4em}{1-shot} & CAT-SAM-T & 95.4 & 90.5 & 93.0 & N/A  & 46.0 & 88.4 &45.4 \\
        & CAT-SAM-A & 94.7 & 90.5 & 92.6 & N/A & 44.9 & 88.0 & 41.4 \\
    \midrule
        \multirow{2}{4em}{16-shot} & CAT-SAM-T & 95.8 & 92.7 & 94.2 & 73.2 & 46.2 & 89.3 & 45.2 \\
        & CAT-SAM-A & 95.5 & 91.5 & 93.5 & 71.4 & 45.7 &89.4 &43.7 \\
    \midrule
        \multirow{2}{4em}{Full-shot} & CAT-SAM-T & 96.3 & 92.6 & 94.4 & 81.7 & 51.4 & 93.0 & 55.0 \\
        & CAT-SAM-A & 96.4 & 92.8 & 94.6 & 82.0 & 52.9 &93.9 &56.0 \\
    \bottomrule
    \end{tabular}
    \label{tab:different-shot-FLS-JSRT}
\end{table}}

\cref{tab:different-shot-FLS-JSRT} shows experimental results. Notably, CAT-SAM-T and CAT-SAM-A improve the mIoU to 93.0\% and 92.6\%, respectively, with only one shot of JSRT training image. The improvements saturate gradually as the number of training samples increases, with marginal gains of 1.4\% and 2.0\%, respectively, when adapted with full-shot target data. These findings underscore the superior few-shot capability of the proposed CAT-SAM and align well with those reported in \cref{tab:SOTA-Fully-Supervised}.
Experiments on FLS~\cite{singh2021marine} show different results. While both CAT-SAM variants exhibit clear adaptation effects, the performance gains for both 16-shot and full-shot adaptation are notably lower as compared with previously evaluated datasets. The discrepancy is largely attributed to two primary factors: 1) the inherent difficulty of FLS with its 11 semantic classes (per-class performances are provided in appendix); and 2) the larger variation of FLS data in shapes and object sizes as compared with previously evaluated data. As for ship instance segmentation using SAR images in HRSID~\cite{wei2020hrsid}, this dataset presents different challenges as ships are very small in size. The results show that CAT-SAM consistently outperforms SAM in this diverse domain, highlighting its efficiency in domain adaptation with large domain discrepancy. These experiments further underscore the need for a more comprehensive adaptation of SAM for exceptionally challenging downstream tasks. We expect further studies along this research direction.

\subsection{Analysis \& Discussion}

\noindent\begin{wrapfigure}[12]{r}{0pt}
    \centering
    \includegraphics[width=0.4\textwidth]{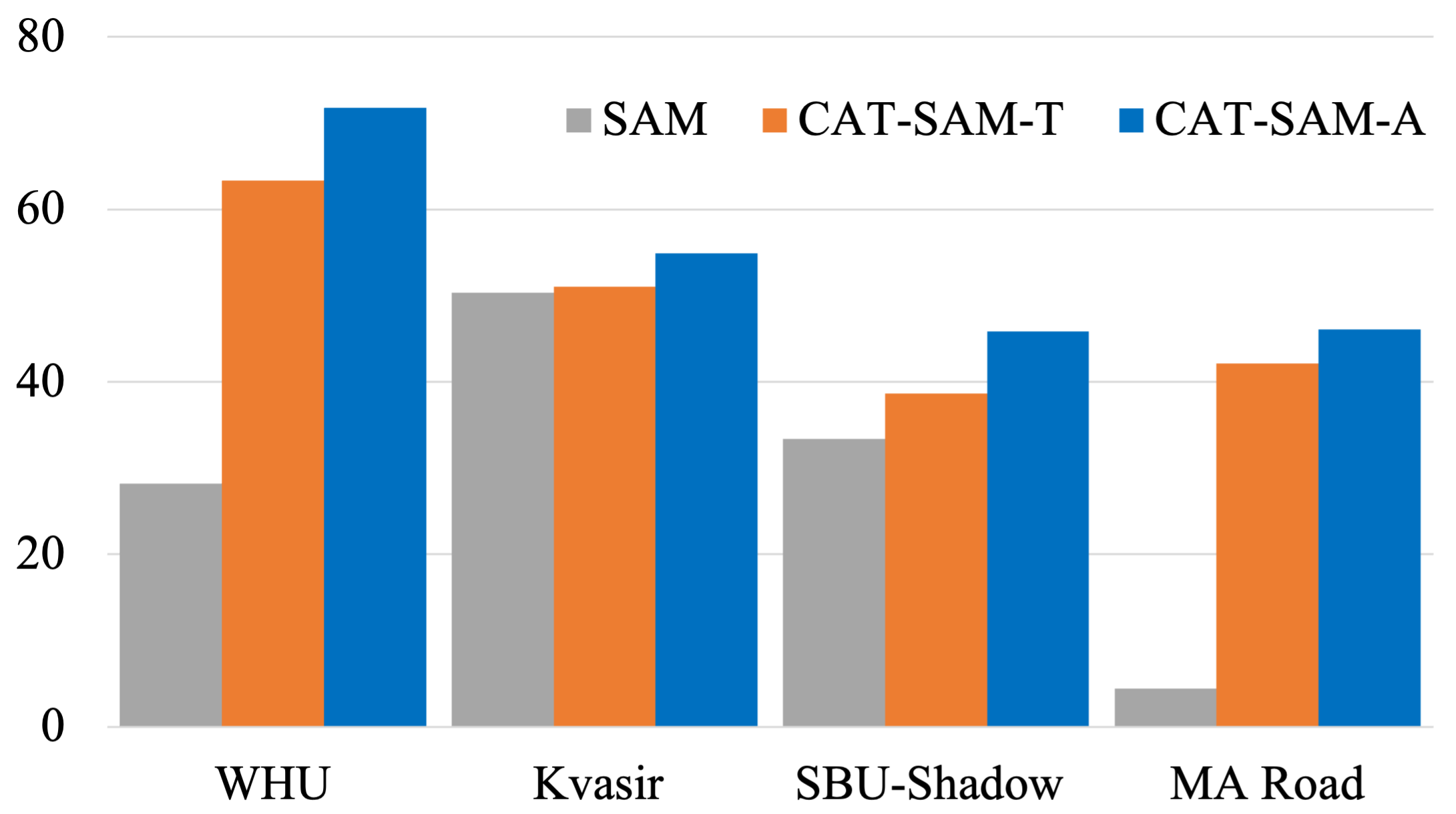}
    \caption{Single point to mask evaluation for one-shot adaptation.}
    \label{fig:SinglePoint}
\end{wrapfigure}
\noindent\textbf{Prompts with Single Point.} SAM provides flexibility in geometric prompts including a single foreground point. However, prompting with a single foreground point can be challenging as the single point could correspond to multiple objects~\cite{kirillov2023segment}. 
Nonetheless, adaptation with few-shot target samples can alleviate this ambiguity by guiding the model to focus on the specific foreground distribution within the annotation space.
We evaluate this by using the same randomly-selected single point as the geometric prompt, and~\cref{fig:SinglePoint} shows the segmentation results for images from WHU, Kvasir, SBU-Shadow, and MA. Roads~\cite{MnihThesis} on road segmentation.

Similar to the experiments using boxes as prompts, CAT-SAM consistently demonstrates superior target segmentation while using a single point as prompt, underscoring its robustness in adaptation while preserving SAM's inherent geometric prompting flexibility. In addition, both CAT-SAM variants achieve relatively less improvement for the dataset Kvasir. This can be largely attributed to the high similarities in color and texture between the polyps and the local gastrointestinal regions in Kvasir, making it challenging for a single point to provide clear geometric guidance.

\begin{figure}[t]
    \centering
    \includegraphics[width=\linewidth]{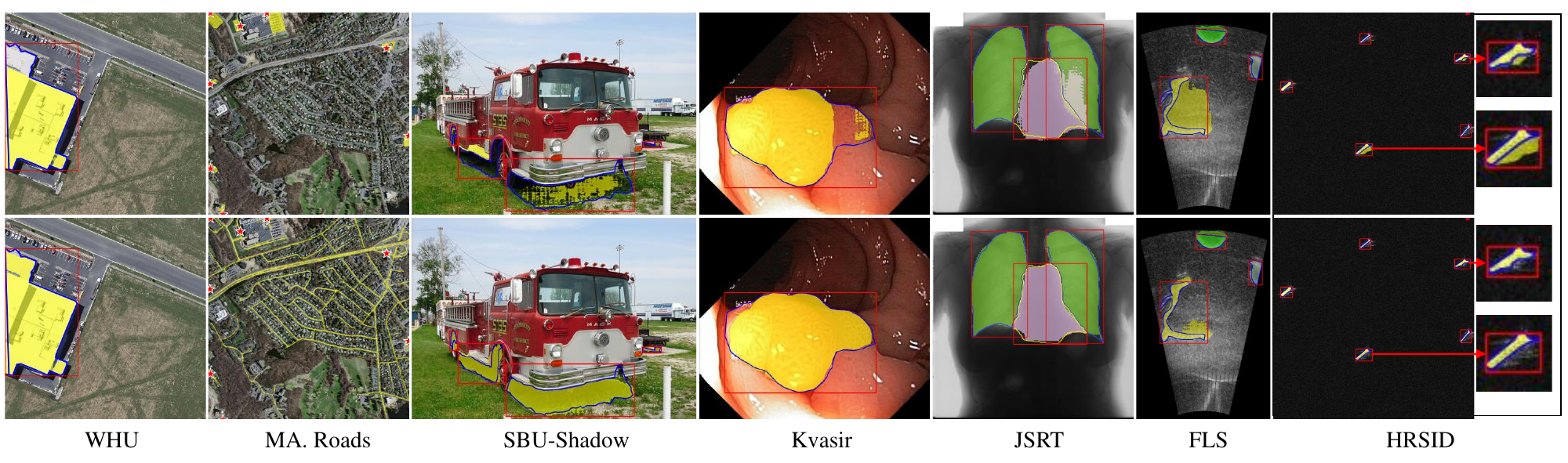}
    \caption{Visual comparisons of SAM~\cite{kirillov2023segment} (top row) and CAT-SAM (bottom row). We illustrate samples from WHU~\cite{ji2018fully} for building segmentation, MA. Roads~\cite{MnihThesis} for road segmentation, SBU-Shadow~\cite{vicente2016large} for shadow segmentation, Kvasir~\cite{pogorelov2017kvasir} for polyp segmentation, JSRT~\cite{shiraishi2000development} for chest organ segmentation (X-ray images), FLS~\cite{singh2021marine} for marine debris segmentation (Sonar images), and HRSID~\cite{wei2020hrsid} for ship instance segmentation (SAR images). CAT-SAM exhibits one-shot adaptation across most datasets, except for 16-shot over FLS. Red boxes and stars denote geometric prompts, colored regions are mask predictions, and lines show the boundary of ground truth segmentation.}
    \label{fig:visual-results}
\end{figure}

\noindent\textbf{Visual Illustrations.}~\cref{fig:visual-results} shows qualitative comparisons between SAM and CAT-SAM across multiple segmentation tasks. These illustrations demonstrate how our proposed CAT-SAM remarkably enhances the mask quality of SAM under the presence of only one-shot target samples. Please refer to the appendix for more visual comparisons.

\section{Conclusion}
We introduce CAT-SAM, a conditional tuning network tailored for few-shot adaptation of SAM. We propose decoder-conditional joint tuning to mitigate the tuning imbalance between SAM's heavyweight image encoder and lightweight mask decoder and facilitate efficient SAM adaptation. To achieve this, we design the prompt bridge structures, enabling interaction when tuning these two network modules. We develop two CAT-SAM variants by integrating the prompt bridge with two representative tuning strategies including prompt tuning and adapter.  Our comprehensive evaluation across diverse scenarios and tasks on 11 segmentation datasets underscores the superior domain adaptive efficiency of both CAT-SAM variants, even in the extremely challenging one-shot scenario.

%
%
\bibliographystyle{splncs04}
\bibliography{arxiv}

\appendix

\section{Implementation Details}

We provide supplementary training details beyond Section 4 of the submitted manuscript. Across all experiments, we employ the AdamW optimizer. The learning rate and weight decay are set at $1 \times 10^{-3}$ and $1 \times 10^{-4}$, respectively. We implement the cosine annealing strategy as the scheduler, with a minimum learning rate of $1 \times 10^{-5}$. Our experiments are conducted using NVIDIA RTX A6000 GPU with 48GB of memory, where a single GPU is utilized for 1-shot adaptation and 4 GPUs for 16-shot and full-shot adaptation. In addition, for the extremely fine-grained experiments conducted on HQ-Seg44k, we adopted the same settings as in \cite{ke2023segment} for fair comparisons. 

For segmentation experiments over WHU, Kvasir, SBU-Shadow, MA. Roads and HRSID, we train 1000 epochs with a batch size of 1 for 1-shot adaptation, 200 epochs with a batch size of 4 for 16-shot adaptation, and 30 epochs for the full-shot adaptation with a batch size of 4. Table 4 at the end of this appendix lists the names of the used few-shot training samples.
We apply random vertical/horizontal flipping and random cropping on each dataset for data augmentation. For input geometrical prompts, we employ the same strategy as HQ-SAM \cite{ke2023segment} by randomly assigning a point, box, or coarse mask for each object. 

We adopted the same training epochs and batch sizes for segmentation experiments on FLS and JSRT, except that we trained 500 epochs on 16-shot adaptation for FLS. For data augmentation, we adopted horizontal flipping only due to the unique geometric structures of the two datasets.

\section{Mask Decoder Details}

\begin{figure}[t]
    \centering
    \includegraphics[width=0.8\linewidth]{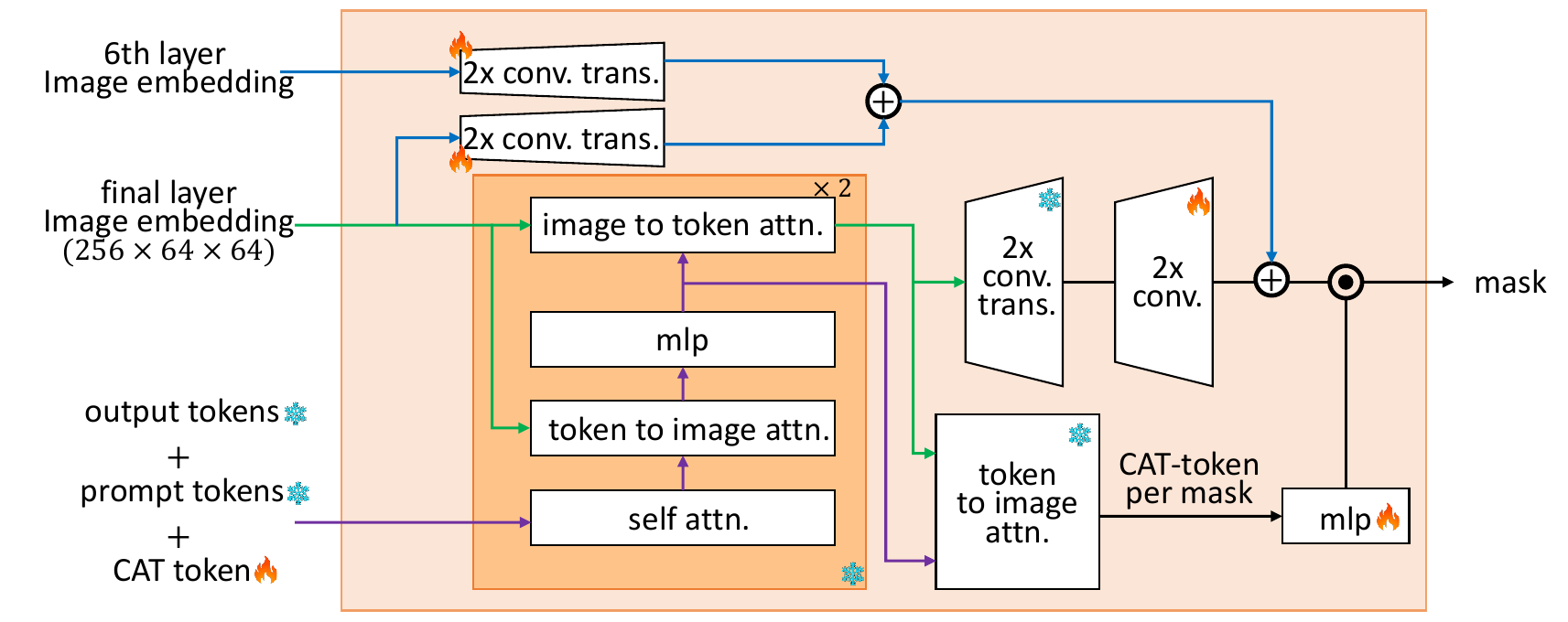}
    \caption{Details of the mask decoder in CAT-SAM. Both CAT-SAM variants share the same decoder tuning structures.}
    \label{fig:decoder}
\end{figure}

As depicted in \cref{fig:decoder}, we freeze the entire decoder of SAM and introduce a learnable \textit{CAT-token} ($1\times256$). This token is concatenated with the output tokens and prompt tokens of the original SAM, forming the input to the mask decoder of CAT-SAM. Following the same pipeline of SAM, the input passes through two decoder layers, and the updated CAT-Token is then employed to generate dynamic weights for a newly introduced three-layer MLP. Additionally, following \cite{ke2023segment}, we upsample and fuse the output image features from the 6th and the last layers of the image encoder. Ultimately, this fused feature is combined with SAM's mask decoder features, and further with the output of the three-layer MLP (via dot product) to produce the target mask.

\setlength{\tabcolsep}{0.5mm}{
\begin{table*}[t]
    \centering
    \tiny
    \caption{Performance comparison of extremely fine-grained segmentation across four intricate structural image datasets. Evaluation of two image encoder backbones demonstrates consistent and substantial segmentation improvements with the two CAT-SAM variants, signifying their robustness and efficacy on SAM adaptation.}
    \begin{tabular}{l|cc|cc|cc|cc|cc|cc}
    \toprule
      \multirow{2}{4em}{Model} & \multicolumn{2}{c|}{Image Encoder} & \multicolumn{2}{c|}{DIS\cite{qin2022highly}} & \multicolumn{2}{c|}{COIFT\cite{liew2021deep}} & \multicolumn{2}{c|}{HRSOD\cite{zeng2019towards}} & \multicolumn{2}{c|}{ThinObject\cite{liew2021deep}} & \multicolumn{2}{c}{Average}\\
      & Backbone & \#Params. & mIoU & mBIoU & mIoU & mBIoU & mIoU & mBIoU & mIoU & mBIoU & mIoU & mBIoU\\
      \midrule
      SAM~\cite{kirillov2023segment}  (baseline) & \multirow{4}{3em}{ViT-L} & \multirow{4}{3em}{308M} & 62.0 & 52.8 & 92.1 & 86.5 & 90.2 &  83.1 &  73.6 & 61.8 & 79.5 & 71.1 \\
      HQ-SAM~\cite{ke2023segment} & & & 78.6 & 70.4 & 94.8 & 90.1 & 93.6  & 86.9 & 89.5 & 79.9 & 89.1 & 81.8 \\
      CAT-SAM-T (ours) & & & 84.0 & 78.1 & 95.6 & 92.0 & 93.4 & 87.6 & 94.0 & 87.9 & 91.7 & 86.4 \\
      CAT-SAM-A (ours) & & & 83.6 & 77.7 & 95.6 & 92.2 & 93.1 & 87.1 & 94.1 & 88.2 & 91.6 & 86.3\\
    \midrule
      SAM~\cite{kirillov2023segment}  (baseline) & \multirow{4}{3em}{ViT-H} & \multirow{4}{3em}{636M} & 57.1 & 49.6 & 91.0 & 86.1 & 87.1 & 80.3 & 68.1 & 58.7 &75.8 & 68.7\\
      HQ-SAM~\cite{ke2023segment} & & & 79.1 & 70.9 & 95.3 & 90.5 & 92.5 & 84.5 & 89.9 & 80.4 &89.2 & 81.6 \\
      CAT-SAM-T (ours) & & & 84.8 & 79.4 & 96.0 & 92.7 & 93.7 & 88.1 & 94.8 & 89.3 & 92.3 & 87.4 \\
      CAT-SAM-A (ours) & & & 84.5 & 79.2 & 96.0 & 92.7 & 93.4 & 87.8 & 94.4 & 88.9 & 92.1 & 87.2\\
    \bottomrule
    \end{tabular}
    \label{tab:fine-seg-backbones}
\end{table*}
}

\section{CAT-SAM across ViT Backbones}
We evaluate CAT-SAM's performance using various backbones of SAM's image encoder. \cref{tab:fine-seg-backbones} presents experiments with ViT-L and ViT-H~\cite{dosovitskiy2020image} image encoders on extremely fine-grained segmentation, following the same setup in Table 6 of our submitted manuscript. The results show that, in comparison to SAM~\cite{kirillov2023segment} and HQ-SAM~\cite{ke2023segment}, both CAT-SAM variants exhibit considerable adaptability when paired with different backbone image encoders. This showcases their superior efficacy in facilitating SAM adaptation. Notably, while ViT-H introduces a substantially higher number of parameters, it only yields marginal performance gains over ViT-L. This observation suggests that scaling up the image encoder further might not give much edge, which aligns well with the findings in \cite{kirillov2023segment}.

\section{More Experimental Results}

\textbf{Per-Class Performance on FLS.} Table \ref{tab:different-shot-FLS} details the per-class performance from Table 7 of the main paper for marine debris segmentation using Sonar images on the FLS dataset~\cite{singh2021marine}. We compare SAM and CAT-SAM variants across 11 classes and their average segmentation results. CAT-SAM variants consistently exhibit superior adaptation effects, with performance improving as the number of samples increases.

\setlength{\tabcolsep}{0.6mm}{
\begin{table}[t]
    \centering
    \scriptsize
    \caption{Few-shot adaptation with varying number of training samples for \textit{non-RGB} datasets, including FLS~\cite{singh2021marine} with Sonar images for marine debris segmentation. 1-shot adaptation unattainable due to the absence of images containing objects from all 11 classes.}
    \begin{tabular}{l|l|cc|c|cccccccc|c}
    \toprule
        \#Samples & Methods & \multicolumn{12}{c}{FLS~\cite{singh2021marine}} \\
    \cmidrule{3-14}
        & & Bott. & Can & Chain & Drin. & Hook & Prop. & Sham. & Stan. & Tire & Valve & Wall & mIoU \\
    \midrule
       None & SAM~\cite{kirillov2023segment} & 76.0 &81.4 &38.4 &78.5 &47.9 &71.6 &81.7 &85.7 &69.3 &54.0 &82.7 &69.7 \\
    \midrule
         \multirow{2}{4em}{1-shot} & CAT-SAM-T &\tikzmark{start}  \\
        & CAT-SAM-A & \\
    \midrule
        \multirow{2}{4em}{16-shot} & CAT-SAM-T &78.1 &76.7 &46.4 &82.3 &49.5 &73.5 &82.8 &85.6 &83.2 &61.8 &85.0 &73.2\tikzmark{end}\\
        & CAT-SAM-A & 72.4 &72.6 &47.9 &78.9 &52.7 &69.6 &81.6 &87.4 &73.8 &62.0 &86.3 &71.4 \\
    \midrule
        \multirow{2}{4em}{Full-shot} & CAT-SAM-T & 86.3 & 87.9 & 66.7 & 83.2 & 67.0 & 79.5 & 87.0 & 88.7 & 90.9 & 70.3 & 91.3 & 81.7 \\
        & CAT-SAM-A & 85.5 &88.1 &67.1 &83.5 &67.6 &80.2 &86.5 &89.8 &91.2 &69.9 &92.1 &82.0\\
    \bottomrule
    \end{tabular}
    \begin{tikzpicture}[overlay, remember picture]
    \coordinate (start adjusted) at ($(start.north west)+(-0.2,0.1)$);
    \coordinate (end adjusted) at ($(end.north east)+(0,0.37)$);
    \draw[line width=0.4pt, black] (start adjusted) -- (end adjusted);
    \end{tikzpicture}
    \label{tab:different-shot-FLS}
\end{table}}

\begin{table}[t]
    \setlength{\tabcolsep}{5.5mm}
    \centering
    \caption{Comparison of adaptive segmentation performance on challenging non-rgb datasets: JSRT~\cite{shiraishi2000development} for chest X-ray segmentation, FLS~\cite{singh2021marine} with Sonar images for marine debris segmentation, and HRSID~\cite{wei2020hrsid} for ship instance segmentation with SAR images. The baseline is SAM without adaptation. We report 1-shot/16-shots/full-shots adaptation results.}
    \label{tab:sota-non-rgb}
    \begin{tabular}{l|ccc}
    \hline
        Model & JSRT (mIoU) & FLS (mIoU) & HRSID (AP) \\
    \hline
        LoRA[18]  & 88.1/94.0/94.2 & -/70.1/80.2 & 41.0/44.7/50.8\\
        HQ-SAM[24] & 89.0/92.1/93.1 & -/70.2/79.5 & 40.1/41.4/48.1\\
        CAT-SAM-T & 93.0/94.2/94.4 & -/73.2/81.7 & 46.0/46.2/51.4\\
        CAT-SAM-A & 92.6/93.5/94.6 & -/71.4/82.0 & 44.9/45.7/52.9\\
    \hline
    \end{tabular}
\end{table}

\noindent \textbf{Comparison with the SoTA for Non-RGB Domains.} In Table \ref{tab:sota-non-rgb}, we benchmark CAT-SAM against the leading methods from each group in Table 4 of the main paper, specifically LoRA \cite{hu2021lora} and HQ-SAM \cite{ke2023segment}, across non-RGB domains including JSRT, FLS, and HRSID. Experiments conducted with 1-shot, 16-shot, and full-shot adaptations consistently demonstrate that CAT-SAM outperforms these methods across all three datasets, particularly with fewer training samples, supporting our conclusions in the main paper.

\noindent \textbf{Scaled Parameters.} We investigate the impact of increasing the number of tokens in the image encoder of CAT-SAM-T (denoted as $P_i$) on adaptation effects. Table \ref{tab:paras} presents the results of 1-shot adaptation with varying token and parameter counts. Doubling the tokens from 1 to 2 results in marginal performance gains across the three datasets. However, further increases in the token count lead to a gradual decline in performance, likely due to the increased parameters requiring more training data for effective tuning. This underscores the effectiveness of using fewer tokens and parameters for few-shot adaptation tasks.

\begin{table}[t]
    \caption{Segmentation of CAT-SAM-T with varing number of tokens in the image encoder for 1-shot adaptation.}
    \label{tab:paras}
    \centering
    \setlength{\tabcolsep}{4mm}
    \begin{tabular}{c|cccccc}
       \hline
       Token number  &  1 & 2 & 4 & 8 & 16 & 32 \\
       \hline
       \#Paras & 3.3M & 5.3M & 9.3M & 17.3M & 33.2M & 65.1M\\
       \hline
       WHU  & 86.8 & 87.4 & 85.7 & 84.3 & 82.4 & 80.2 \\
       Kvasir & 83.4 & 85.2 & 81.4 & 81.3 & 79.1 & 73.9\\
       SBU & 78.0 & 79.0 & 76.6 & 75.3 & 70.3 & 67.9\\
       \hline
    \end{tabular}
\end{table}

\section{More Visual Illustrations}

\begin{figure}[t]
    \centering
    \includegraphics[width=\linewidth]{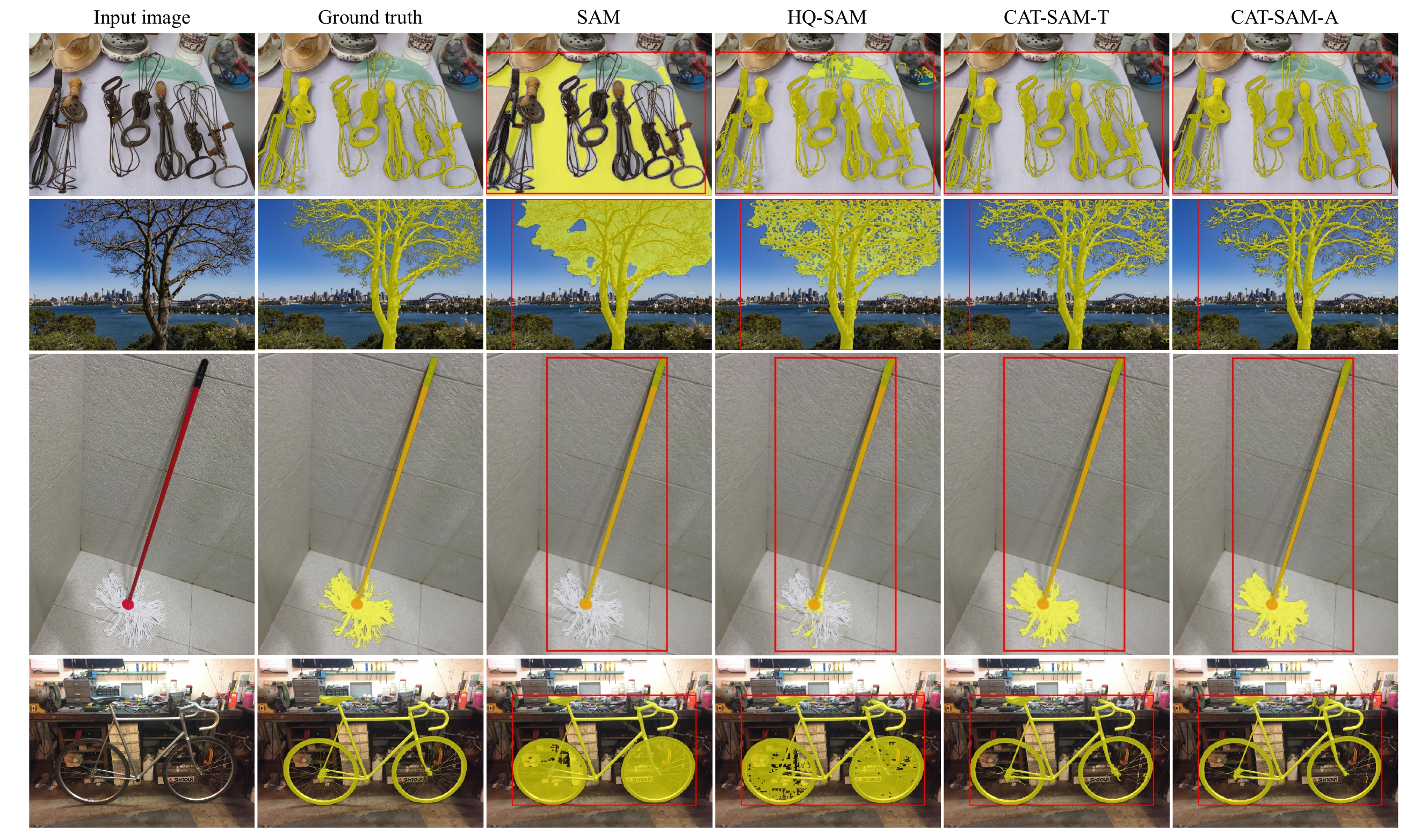}
    \caption{Visual comparisons of SAM, HQ-SAM, and the proposed CAT-SAM for extremely fine-grained segmentation.}
    \label{fig:finegrained-seg}
\end{figure}

We present additional visual illustrations showcasing the efficacy of CAT-SAM on SAM adaptation. \cref{fig:finegrained-seg} shows qualitative experiments corresponding to Table~6 in our submitted manuscript, illustrating the extremely fine-grained segmentation of objects with intricate structures. These examples show SAM's limitations clearly in generating high-quality segmentations for such objects. 
Though HQ-SAM achieves noticeable improvements, its segmentation remains clearly different from the ground truth. As a comparison, the two CAT-SAM variants exhibit  much better accuracy in segmenting these challenging samples.
The superior segmentation performance of the two CAT-SAM variants underlines their effectiveness in adapting to complex structural images.

\begin{figure}[t]
    \centering
    \includegraphics[width=\linewidth]{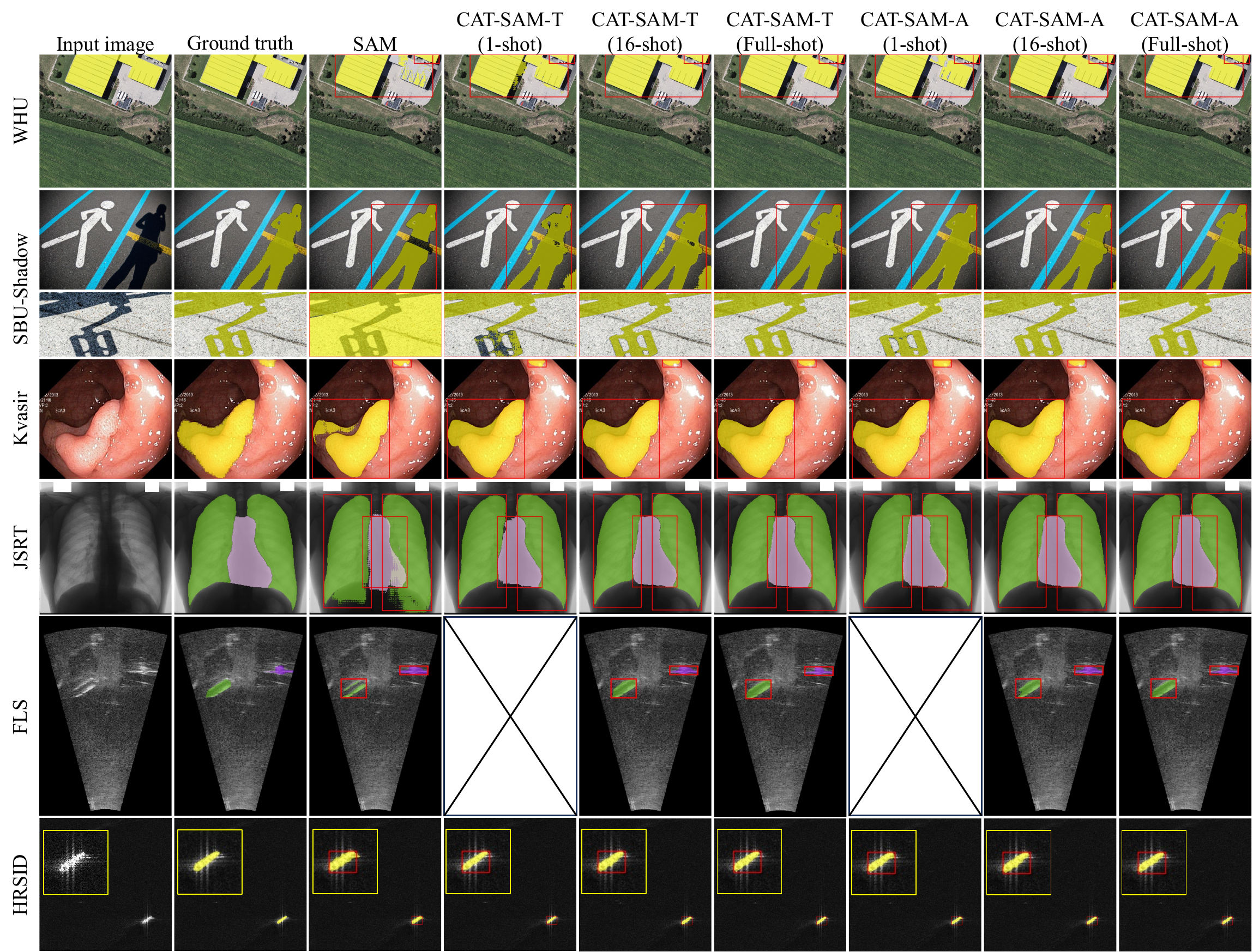}
    \caption{Visual comparisons of SAM and adapted CAT-SAM with different target shots. 1-shot adaptation over FLS is unattainable due to the absence of images containing objects from all 11 classes.}
    \label{fig:multishot-seg}
\end{figure}

\cref{fig:multishot-seg} shows qualitative experiments corresponding to Tables 5, 7, and 8 of our submitted manuscript. The visual illustrations consistently corroborate the quantitative findings, demonstrating that employing more target samples consistently enhances segmentation performance across diverse and challenging RGB and non-RGB datasets. Remarkably, even with just one-shot target samples, both variants of CAT-SAM achieve remarkably high-quality target segmentation. This underscores the robustness and effectiveness of CAT-SAM in handling diverse target samples, showcasing its adaptability and generalization capabilities.

\section{Robustness Analysis}

We examine the robustness of CAT-SAM by evaluating it with diverse target samples under the challenging 1-shot adaptation setup. Specifically, we randomly select three distinct training images from each of the WHU, Kvasir, and SBU-Shadow datasets and train three 1-shot adaptation models, each following the same setup as detailed in Table 4 of the manuscript. Subsequently, we report the averaged segmentation performance over the testing set for the three models, along with upper and lower deviations. \cref{tab:multiple-run} summarizes the experimental results. It is evident that across all three datasets, both variants of CAT-SAM consistently demonstrate notable 1-shot adaptation effects with varying training samples, demonstrating the superior robustness of our CAT-SAM.

\setlength{\tabcolsep}{4.8mm}{
\begin{table}[t]
    \scriptsize
    \centering
    \caption{1-shot adaption of CAT-SAM variants across challenging true color image datasets. Both CAT-SAM variants achieve superior and robust adaptation results when trained with different training samples.}
    \begin{tabular}{l|ccc}
    \toprule
       Methods & WHU & Kvasir & SBU-Shadow \\
    \midrule
      SAM~\cite{kirillov2023segment} (baseline) & 43.5 & 79.0 & 62.4 \\
    \midrule
      CAT-SAM-T (Ours) & 87.3±0.5 & 83.2±0.5 & 79.0±1.0 \\
      CAT-SAM-A (Ours) & 88.5±0.4 & 85.7±0.5 & 84.1±2.2 \\
    \bottomrule
    \end{tabular}
    \label{tab:multiple-run}
\end{table}
}

\section{Failure Case Analysis}
\begin{figure*}[!h]
    \centering
    \includegraphics[width=\linewidth]{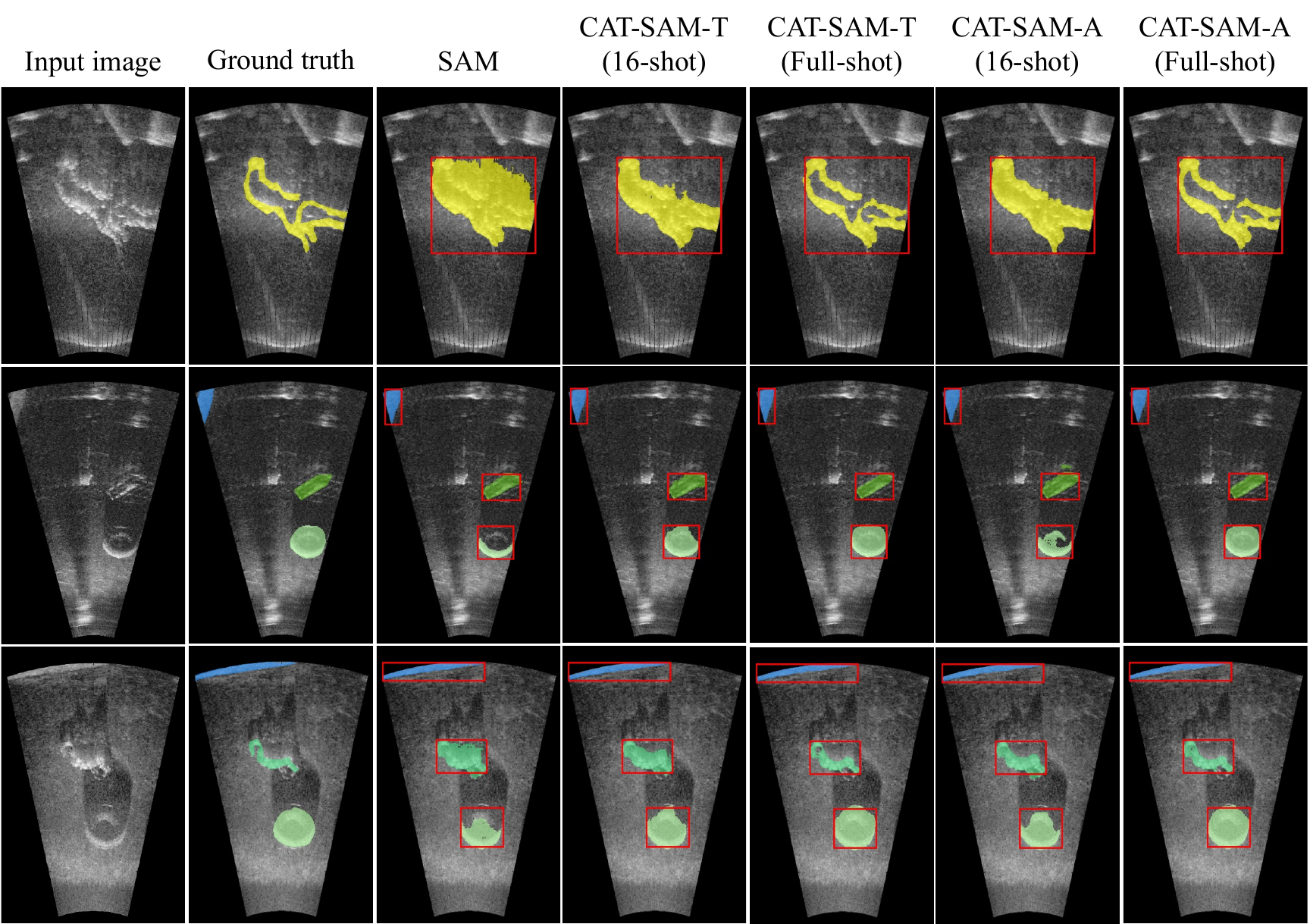}
    \caption{Visual comparisons of SAM and adapted CAT-SAM with different target shots over the FLS dataset.}
    \label{fig:FLS}
\end{figure*}

\cref{fig:FLS} visualizes SAM and CAT-SAM segmentation while facing challenging samples, where the experiments are conducted with 16-shot and full-shot on the FLS dataset. 
Though both CAT-SAM variants demonstrate great improvements over SAM under the 16-shot setup, they still exhibit clear disparities from the ground-truth segmentation. In contrast, full-shot adaptation achieves significantly better segmentation over these samples.  This is consistent with the points we stated in the limitations part of the manuscript: more research is needed for accurate and reliable segmentation especially while handling data-scarcity scenarios in challenging real-world tasks.

\section{Limitations} 
Although CAT-SAM introduces a limited number of additional parameters, it remains computationally demanding and cannot achieve real-time speed due to its reliance on SAM, necessitating substantial GPU resources and limiting its applicability for tasks such as video processing.
Additionally, its adaptation performance, particularly in highly complex domains like Sonar images with multiple classes, needs further improvements for various real-world applications especially in data-scarce scenarios.

\section{More Experimental Details}
\cref{tab:filenames} lists the file names of training samples employed in both 1-shot and 16-shot adaptation experiments as described in the manuscript.

\begin{longtable}{l|l}
    \caption{File names for 1-shot and 16-shot training.} 
    \label{tab:filenames} \\
        \toprule
        \textbf{Dataset} & \textbf{File Names} \\
        \midrule
         \multicolumn{2}{c}{\textbf{1-shot}} \\
        \midrule
        Kvasir & cju0roawvklrq0799vmjorwfv\\
        WHU & 2432\\
        SBU & lssd1409\\
        JSRT & case164 \\
        M Road & 23879065\_15 \\
        HRSID & P0003\_600\_1400\_9189\_9989 \\
        \midrule
        \multicolumn{2}{c}{\textbf{16-shot}} \\
        \midrule
        Kvasir & cju7dn24o296i09871qfxb8s2, \\ 
        & cju83mki1jv5w0817kubxm31r, \\
        & cju6vucxvvlda0755j7msqnya, \\
        & cju43kj2pm34f0850l28ahpni, \\
        & cju2x7vw87mu30878hye2ca0m, \\
        & cju5ddda9bkkt0850enzwatb1, \\
        & cju77vvcwzcm50850lzoykuva, \\
        & cju3128yi0rpu0988o4oo5n8n, \\
        & cju8b7aqtr4a00987coba14b7, \\
        & cjz14qsk2wci60794un9ozwmw, \\
        & cju0s690hkp960855tjuaqvv0, \\
        & cju888fr7nveu0818r9uwtiit, \\
        & cju8b542nr81x0871uxnkm9ih, \\
        & cju2qtee81yd708787bsjr75d, \\
        & cju83h9ysjwe808716nt35oah, \\
        & cju0roawvklrq0799vmjorwfv \\
        \midrule
        WHU & 168, 515, 807, 1176, 1552, 1901, 2432, 2679, \\
        & 3130, 3486, 3678, 3939, 4263, 4335, 4640, 4725 \\
        \midrule
        SBU & lssd3356, lssd1409, lssd1090, lssd3732, \\
        & lssd2011, lssd1901, lssd1821, lssd1512, \\
        & lssd3713, lssd1376, lssd3493, lssd3729, \\
        & lssd610, lssd3008, lssd1319, lssd1109 \\
        \midrule
        JSRT & case007, case009, case023, case027, \\
        & case029, case036, case058, case063, \\
        & case071, case109, case140, case152, \\
        & case164, case174, case189, case190 \\
        \midrule
        M Road & 10528660\_15, 10528735\_15, 15928885\_15, 20128945\_15, \\
        & 20578960\_15, 21928990\_15, 22378945\_15, 22979005\_15, \\
        & 23278960\_15, 23428960\_15, 23728540\_15, 23879065\_15, \\
        & 24178525\_15, 24628855\_15, 24629290\_15, 26128780\_15 \\
        \midrule
        FLS & marine-debris-aris3k-53, marine-debris-aris3k-295, \\
        & marine-debris-aris3k-298, marine-debris-aris3k-322, \\
        & marine-debris-aris3k-365, marine-debris-aris3k-391, \\
        & marine-debris-aris3k-744, marine-debris-aris3k-1039, \\
        & marine-debris-aris3k-1307, marine-debris-aris3k-1324, \\
        & marine-debris-aris3k-1359, marine-debris-aris3k-1393, \\
        & marine-debris-aris3k-1457, marine-debris-aris3k-1774, \\
        & marine-debris-aris3k-1816, marine-debris-aris3k-1817 \\
        \midrule
        HRSID & P0081\_2400\_3200\_4200\_5000, \\
        & P0012\_5400\_6200\_7200\_8000, \\
        & P0105\_3600\_4400\_16200\_17000, \\
        & P0010\_2400\_3200\_4800\_5600, \\
        & P0076\_0\_800\_4200\_5000, \\
        & P0064\_1800\_2600\_5400\_6200, \\
        & P0060\_1200\_2000\_6000\_6800, \\
        & P0056\_1200\_2000\_5400\_6200, \\
        & P0067\_600\_1400\_6000\_6800, \\
        & P0002\_1200\_2000\_8400\_9200, \\
        & P0027\_1800\_2600\_8289\_9089, \\
        & P0109\_3000\_3800\_7200\_8000, \\
        & P0108\_2400\_3200\_7200\_8000, \\
        & P0109\_3000\_3800\_9600\_10400, \\
        & P0105\_3600\_4400\_15600\_16400, \\
        & P0028\_600\_1400\_7200\_8000 \\
    \bottomrule
\end{longtable}

\end{document}